\documentclass{article}

\usepackage{enumitem}
\usepackage[most]{tcolorbox}
\usepackage{amssymb}
\usepackage[ruled,linesnumbered]{algorithm2e}
\usepackage{nicematrix}
\usepackage{makecell}
\usepackage{amsmath}
\usepackage{graphicx}
\usepackage{wrapfig}

\usepackage[main, final]{neurips_2026}

\usepackage[utf8]{inputenc} 
\usepackage[T1]{fontenc}    
\usepackage{hyperref}       
\usepackage{url}            
\usepackage{booktabs}       
\usepackage{amsfonts}       
\usepackage{nicefrac}       
\usepackage{microtype}      
\usepackage{xcolor}         
\usepackage{booktabs, multirow}
\usepackage{soul}

\usepackage{graphicx}
\usepackage[table]{xcolor}
\usepackage{multirow}
\usepackage{booktabs}
\definecolor{glmcolor}{HTML}{EEEDFE}   
\definecolor{minimaxcolor}{HTML}{E6F1FB} 
\definecolor{mimocolor}{HTML}{FFF9C4}  

\newcommand{\method}{\textsc{{SelfCompact}}}

\definecolor{darkblue}{rgb}{0, 0, 0.5}
\hypersetup{colorlinks=true, citecolor=darkblue, linkcolor=darkblue, urlcolor=darkblue}

\title{Self-Compacting Language Model Agents}

\author{%
  Tianjian Li$^{\spadesuit}$~~~Jingyu Zhang$^{\spadesuit}$~~~William Jurayj$^{\spadesuit}$~~~Xi Wang$^{\spadesuit}$~~~Chuanyang Jin$^{\spadesuit}$ \\
  \textbf{Mehrdad Farajtabar$^{\heartsuit}$~~~Eric Nalisnick$^{\spadesuit}$~~~Daniel Khashabi$^{\spadesuit}$} \\
  \\
  $^{\spadesuit}$Johns Hopkins University~~$^{\heartsuit}$Apple \\
  \texttt{\{tli104, danielk\}@jhu.edu} \\
}

\begin{document}

\maketitle

\begin{abstract}
Long agent traces composed of chains of thought and tool calls accumulate stale content that anchor subsequent generations, and eventually outgrow the context window. Existing scaffolds mitigate it with
\textit{fixed-interval} compaction triggered at a token threshold. Such triggers pay no heed to trajectory structure, risking
discard of partial results mid-derivation or mid-search. We propose
\method{}, a scaffold that allows the model itself to decide \textit{when} and \textit{how} to compact. Specifically, it
pairs two inference-time elements: 
(i) a
\emph{compaction tool} the model invokes to summarize the accumulated context, and  
(ii) a lightweight
\emph{rubric} specifying when to fire (a sub-task has resolved, or
the trajectory is converging) and when to suppress (mid-derivation,
or when stuck). Both are needed. The tool alone is unevenly used
across open-weight models, often invoked at unhelpful moments or not
at all; the rubric alone cannot act. Together, they elicit effective
adaptive compaction without any fine-tuning or external supervision. We present empirical results on six benchmarks (competitive math
and agentic search) and seven models. Our results show that
\method{} matches or exceeds fixed-interval summarization at a fraction of the token cost, improving over a no-summarization baseline by up to 18.1 points on math and 5--9 points on agentic search at 30--70\% lower
per-question cost. Our results expose a meta-cognitive gap: although unprompted models cannot reliably tell when their own context is rotting, a lightweight rubric closes this gap, reframing when to compact as a capability that scaffolds can supply without training\footnote{Code is available at \href{https://github.com/tianjianl/selfcompact}{https://github.com/tianjianl/selfcompact}.}. 

\end{abstract}

\section{Introduction}
\label{sec:intro}

We are chasing after harder problems over longer horizons \citep{metr-2026-time-horizon-1-1}, and consequently, the trajectories LMs generate to solve them keep growing. Reasoning models are now able to spend tens of thousands of tokens deliberating on a single competition math question: Qwen3.5 produces 81k tokens \citep{qwen3.5}, Kimi-K2.5 produces 96k \citep{kimiteam2026kimik25visualagentic}. Agentic systems extend further, orchestrating search results \citep{wei2025browsecompsimplechallengingbenchmark}, code execution outputs \citep{jimenez2024swebench}, and intermediate plans \citep{novikov2025alphaevolvecodingagentscientific} across hundreds of turns. The bet that more thinking and interactions yields better answers has paid off, but long traces carry a hidden cost.

As the trace grows, it accumulates junk --- a flawed case split made early, a search result the model has moved past, a candidate program that led nowhere. These leftovers do not just sit there; they anchor everything that follows \citep{laban2026llms}. A model that solves a problem from a clean start often fails when fed back its own flawed reasoning. This phenomenon is known as \emph{context rot} \citep{hong2025context, cheng2026contextualdrag}. Existing systems try to manage it with rigid rules: compacting the trajectory when a token threshold is met \citep{research2026composer2technicalreport}, or delegating the burden of identifying context rot to the user via \texttt{/compact} \citep{anthropic2025claudecode}.

\begin{figure}[t!]
    \centering
    \includegraphics[width=\linewidth]{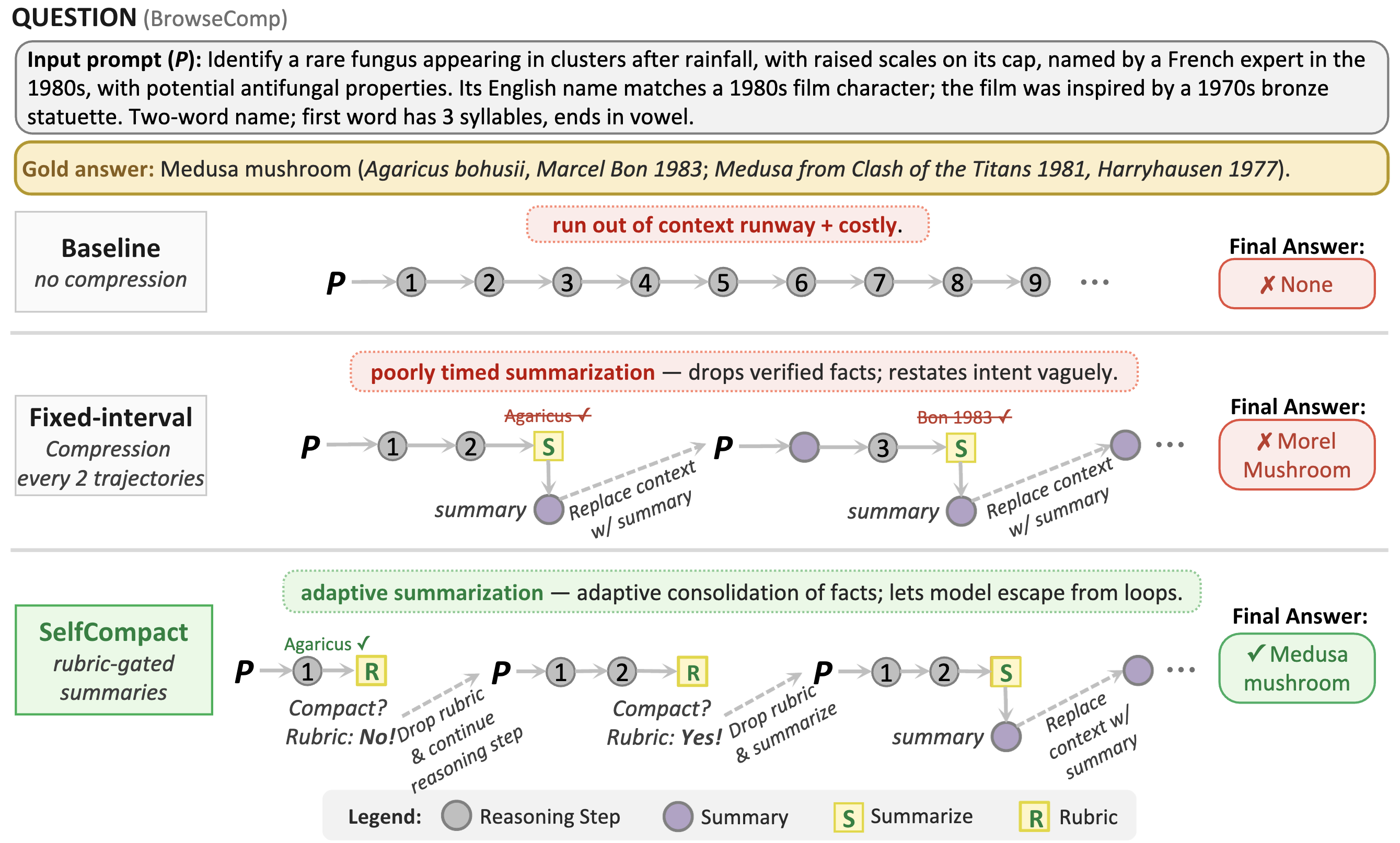}
    \caption{\textbf{Comparison of trajectory-compression strategies on a hard
BrowseComp question.} The gold answer requires verifying four facts
(\emph{Agaricus}, \emph{Bon 1983}, \emph{Clash 1981}, \emph{Harryhausen})
before composing \emph{Medusa mushroom}. \textbf{Baseline} (no compression)
burns its budget on an unproductive monologue and emits no answer.
\textbf{Fixed-interval} compression fires every two search trajectories regardless
of reasoning state; the poorly-timed summary wipes verified facts
mid-reasoning, leaving the model to guess (\emph{Morel Mushroom}).
\textbf{\method{} (Ours)} gates compression on a rubric over closed
reasoning units, so each summary fires immediately after a verified fact---all
four facts persist and the model returns the correct answer.}
\label{fig:teaser}
\end{figure}

  Today, this rigidity is the rule. Industrial deployed systems either wait until
  fixed token intervals to invoke compaction \citep{research2026composer2technicalreport}; recent academic scaffolds that compact context on reasoning traces \citep{wu2026reasoningcachecontinualimprovement, yan2026inftythink} and deep research systems \citep{chen2026iterresearch}, all rely on a token threshold to trigger compaction. For example, \citet{liu2025webexplorer} triggers compaction when ``token usage exceeds 30\% of the maximum context.''\footnote{\url{https://huggingface.co/MiniMaxAI/MiniMax-M2.5}}  
  But a fixed threshold knows nothing about what the model is doing when it invokes summarization. It cannot distinguish a model mid-derivation from one that has just resolved a sub-problem, mid-search from one converging on an answer. The cost is asymmetric: a well-timed call discards stale work, but a poorly timed call discards the partial result the model needed to continue. These scaffolds help, but they leave substantial headroom. The right time to compact heavily depends on what the model is doing, not just on how many tokens have passed.  Figure~\ref{fig:teaser} shows this concretely: fixed-interval
compression wipes four already-verified facts mid-trajectory, and the
model falls back to a guess. In this work, we ask the following research question:
  
\begin{tcolorbox}[enhanced,
  colback=gray!8, colframe=gray!90,
  boxrule=0.4pt, arc=2pt,
  left=2pt, right=2pt, top=4pt, bottom=4pt,
  drop fuzzy shadow=black!20, width=1.0\linewidth, center]
  \begin{center}
    \textit{Can the LM agent itself }recognize\textit{ its own context rot without training, and compact accordingly?}
  \end{center}
\end{tcolorbox}

To answer the research question, we hand the decision of when to compact back to the LM agent itself. We propose \method{}, a scaffold that pairs two inference-time elements: an inline \emph{compaction tool} the model invokes itself, and a lightweight \emph{rubric} specifying when to fire (a sub-task has resolved, or the trajectory is converging) and when to suppress (mid-derivation, or when stuck). The model emits \texttt{<summarize>} when it wants to compact, the scaffold runs the summarization, and generation resumes from the summarized prefix. Both the rubric judge and summarizer is the same model: no external verifier, no fine-tuning.

Both elements are needed. Across seven open-weight models, four \texttt{Qwen} models on competition math and three deployed search agents (\texttt{MiniMax-M2.5}, \texttt{GLM-4.7-Flash}, \texttt{MiMo-V2-Flash}), ablating the rubric exposes uneven tool use: some models call the tool at unhelpful moments, others not at all. With the rubric in place, \method{} elicits effective adaptive compaction without any fine-tuning, matching or exceeding fixed-interval summarization at a fraction of the token cost: on competition math (IMO-Answerbench, HMMT Nov 25 / Feb 26), it improves over a no-compaction baseline by up to 18.1 points; on agentic search (BrowseComp, BrowseComp-Plus, DeepSearch QA), it adds 5--9 points at 30--70\% lower per-question cost. 

To sum up, our contribution is three-fold:
\begin{itemize}[leftmargin=*]
\item  We propose \method{}, a novel scaffold that exposes summarization as an inline tool paired with a lightweight rubric that tells the model when to invoke it. No fine-tuning required.
\item  Across seven open-weight models on competition math (four reasoning models) and agentic search (three deployed agents), \method{} matches or exceeds fixed-interval summarization at substantially lower token cost.
\item Ablations find that the rubric is crucial for effective self-compaction: offering the tool alone yields uneven behavior, while a paragraph of guidance closes the gap. Our findings imply that knowing when to perform compaction can be supplied as a meta-cognitive capability at inference time, rather than baked into weights.
\end{itemize}

To the best of our knowledge, our work is the first to introduce and evaluate rubric-based adaptive compaction timing in a training-free setting.

\section{Preliminaries}

\paragraph{Context management in long-horizon agents.} LM agents on long-horizon tasks accumulate tokens in a single rolling context: prior thoughts, tool calls, and observations. A single coding-agent session on SWE-bench-style tasks routinely spans millions of tokens and well over a hundred turns. e.g., Qwen3-Coder-Next averages 8M tokens and 154 turns per problem on SWE-rebench~\citep{badertdinov2026swerebench}. Deep-research agents on BrowseComp scale to hundreds of tool calls per query~\citep{wei2025browsecompsimplechallengingbenchmark, miromindteam2026mirothinkerpushingperformanceboundaries}. Even single-shot reasoning is no longer short: Kimi-K2.5 \citep{kimiteam2026kimik25visualagentic} generates up to 96k thinking tokens on a single competition math problem. As horizons and task difficulty continue to scale, this accumulated context is not merely expensive but actively harmful: model performance degrades sharply with length, a phenomenon known as \emph{context rot}~\citep{hong2025context}. Compaction is therefore a prerequisite for sustained long-horizon capability.

\paragraph{Current strategies.} Today's agents compact via two
content-agnostic heuristics. \emph{Reactive} compaction triggers only when
the rolling context approaches the model's token budget, treating
compaction purely as overflow prevention~\citep{anthropic2025claudecode}.
\emph{Periodic} compaction fires on a \emph{fixed interval}---every $k$
turns or every $k$ tokens---regardless of what the trajectory contains
\citep{research2026composer2technicalreport,
wu2026reasoningcachecontinualimprovement}. Both ignore the state of the
trajectory, and fail in opposite directions: reactive compaction waits
until the context is already saturated with stale or erroneous tokens that
have been polluting generation for many steps, while periodic compaction
discards context indiscriminately, often summarizing in the middle of an
active subgoal and erasing information the model still needs.
Figure~\ref{fig:teaser} illustrates how a fixed-interval
summary firing mid-reasoning on a BrowseComp trajectory can wipe four
already-verified facts, and the model is left to guess.

\section{Our Approach: \method{}}


\SetKwInOut{Note}{Notation}

\begin{algorithm}[t]
\caption{\method{}: rubric-gated inference-time context compaction.}
\label{alg:selfcompact}
\KwIn{prompt $x$; model $\pi$; probe interval $N$; step budget $T$;
      rubric prompt $P_R$; summarizer prompt $P_S$.}
\Note{$\mathcal{C} \leftarrow \mathcal{C} \circ m$ appends message $m$ to context $\mathcal{C}$;
        $z \sim \pi(\cdot\mid\mathcal{C})$ samples response $z$ from $\pi$ given $\mathcal{C}$;
        $\text{pop}(\mathcal{C}, m)$ pop (removes) message $m$ from $\mathcal{C}$ (assuming that it's the last item in $\mathcal{C}$).}
$\mathcal{C} \gets x$\tcp*{Overall context and its KV cache} 
\For{$t = 1, \dots, T$}{
    $y_t \sim \pi(\cdot \mid \mathcal{C})$;\ \ $\mathcal{C} \leftarrow \mathcal{C} \circ y_t$\;
    \lIf{$y_t$ is a final answer}{\Return $y_t$}
    \If{$t \bmod N = 0$}{
        $\mathcal{C} \leftarrow \mathcal{C} \circ P_R$\tcp*{rubric probe; KV prefix reused}
        $r_t \sim \pi(\cdot \mid \mathcal{C}),\ r_t \in \{\textsc{compress},\textsc{continue}\}$;\ \ $\mathcal{C} \leftarrow \mathcal{C} \circ r_t$\;
        \eIf{$r_t = \textsc{compress}$}{
            $\mathcal{C} \leftarrow \mathcal{C} \circ P_S$\tcp*{KV prefix still reused}
            $\tilde{y} \sim \pi(\cdot \mid \mathcal{C})$\tcp*{summarize trajectory}
            $\mathcal{C} \gets x \circ \tilde{y}$\tcp*{hard reset; resume from summary}
        }{
            $\text{pop}(\mathcal{C}, r_t)$;\ \ $\text{pop}(\mathcal{C}, P_R)$\tcp*{revert; trajectory unchanged}
        }
    }
}
\Return $y_T$\;
\end{algorithm}

\paragraph{Setup.} Given a natural language prompt $x$, a language model $\pi(\cdot \mid x)$ generates a continuation $y = (y_1, y_2, \ldots)$ autoregressively, with each token conditioned on the prompt and all preceding tokens. We equip the language model with a summarization tool $\mathcal{S}: (x, y_{1:t}) \rightarrow \tilde{y}$, which takes the original prompt $x$ and a (possibly partial) continuation $y_{1:t}$ and produces a condensed version $\tilde{y}$. Generation then resumes from the new context $(x, \tilde{y})$. Crucially, the summarizer is the same model $\pi$---no external verifier or auxiliary model is required.

\paragraph{\method{}.} \method{} pairs two inference-time elements: (i) the summarization tool $\mathcal{S}$ exposed to the model, and (ii) a lightweight \emph{rubric} that decides, at periodic probe intervals, whether $\mathcal{S}$ should fire. Both elements are needed: exposing the tool alone---letting $\pi$ emit \texttt{<summarize>} whenever it wants---yields uneven, often poorly-timed firing across models: some call the tool reflexively at unhelpful moments, others rarely at all. The rubric closes that gap by translating ``a sub-task has resolved'' and ``the trajectory is not stuck mid-derivation'' into concrete, cite-able conditions $\pi$ can verify against the trajectory in front of it. \method{} runs on a single inference engine with no fine-tuning.

\paragraph{Summarizer design.} To maximize KV-cache reuse, we implement $\mathcal{S}$ by appending a summarizer prompt (Box~\ref{box:math_summarizer} for Math; Box~\ref{box:summ} for Search Agents) as a user-role message: the cache of $y_{1:t}$ at trigger time is preserved across the call,
so $\mathcal{S}$ pays prefill only on the appended instruction, not on the full trajectory. A saving we quantify in
the cost analysis below. The scaffold appends the instruction to
$(x, y_{1:t})$, $\pi$ generates the summary $\tilde{y}$, and $\tilde{y}$ then
replaces $y_{1:t}$; decoding resumes from the original question and the summarized content $(x, \tilde{y})$. 

\paragraph{Triggering the summarizer.} At intervals
$t \in \{N, 2N, 3N, \ldots\}$, the scaffold appends a rubric prompt $P_R$ as a user-role message to
$(x, y_{1:t})$ and samples from $\pi(x, y_{1:t},P_R)$ a binary verdict $r_t \in
\{\textsc{compress}, \textsc{continue}\}$. These intervals are measured in tokens of the reasoning trace (math) or in tool calls (agentic search).

As with $\mathcal{S}$, the probe is
appended rather than substituted so that the KV cache of $y_{1:t}$ is
preserved across the call, keeping the judge near-free relative to the
trajectory length. The probe enumerates a small set of conditions, each
requiring verbatim evidence quoted from the trajectory, so $\pi$ can check
them locally. The rubric is task-specific: the conditions for competition
math differ from those for agentic search, and we defer the exact wording to
Appendix~\ref{appendix:math_setup} (math) and
Appendix~\ref{appendix:search_setup} (search). On a \textsc{compress} verdict,
the summarizer fires:
\[
\tilde{y} \leftarrow \mathcal{S}(x, y_{1:t}) \quad \text{whenever } t \equiv 0 \pmod{N} \text{ and } r_t = \textsc{compress};
\]
otherwise generation resumes from $(x, y_{1:t})$ unchanged.

\paragraph{Workflow.} Our experiments compare three regimes that share the same generation loop but differ in how the transcript is compressed: \textbf{no compaction} (the trajectory accumulates untruncated), \textbf{fixed-interval compaction} (summarize-and-replace at every probe interval, and \method{} (summarize-and-replace only when the rubric returns \textsc{compress}, otherwise continue). On agentic search this yields a \emph{search--judge--summarize--search} loop, with the agent continuing against $\tilde{y}$ rather than the full history after each compression. 

\paragraph{Cost analysis.} Define $L = |y_{1:t}|$ as the pre-compression
trajectory length and $\ell = |\tilde{y}|$ for the summary length, both in
tokens. \method{} adds at most two LLM calls per probe interval: the rubric
probe always, the summarizer only on a \textsc{compress} verdict. 
Both append to $(x, y_{1:t})$, so the KV cache of $y_{1:t}$ is reused
across the call and the $\mathcal{O}(L^2)$ re-prefill that a naive
re-encode would incur is avoided. The rubric probe is essentially free: a
short verdict $r_t$ generated on top of the cached prefix. The summarizer
is the only real overhead: $\mathcal{O}(L\ell)$ to generate $\tilde{y}$
plus a one-time $\mathcal{O}(\ell^2)$ prefill of the new $(x, \tilde{y})$
prefix. Empirically the trade is favorable: on agentic search
\method{} adds $5$--$9$ accuracy points at $30$--$70\%$ lower per-question
token cost than the no-compaction baseline (\S\ref{subsec:agentic-search});
on competition math it leads on $11$ of $12$ benchmark/model cells under a
token budget matched to fixed-interval summarization
(\S\ref{subsec:multi-turn-math}). Appendix~\ref{appendix:cost} states
the per-question cost formula in full, defines how
$(N_{\text{prefill}}, N_{\text{cached}}, N_{\text{out}})$ are
accumulated from each trajectory, lists the OpenRouter prices, and
works out the cached-vs-prefill break-even (compaction wins iff $L/\ell
> 10$, which the search summarizer reaches at $20$--$80\times$).

\section{Experiments}

\subsection{Multi-turn Reasoning in Competition Math}
\label{subsec:multi-turn-math}

\paragraph{Setup.} We evaluate 4 models of different sizes in the Qwen family --- 2 instruct models \texttt{Qwen3-4B-Instruct-2507} and \texttt{Qwen3-30B-A3B-Instruct-2507}, and 2 Qwen3.5 \citep{qwen3.5}: \texttt{Qwen3.5-4B} and \texttt{Qwen3.5-9B}, on 3 benchmarks: IMO-Answerbench \citep{luong-etal-2025-towards}, HMMT Nov 2025 and HMMT Feb 2026 \citep{balunovic2025matharena}. We generate 16 responses for each question in parallel and report the mean across samples. We use vLLM \citep{kwon2023efficient} for inference and the \href{https://github.com/huggingface/Math-Verify}{\texttt{math\_verify}} package for parsing and evaluation.

We compare \method~ against fixed interval summary, where summarization is always triggered for every 16k tokens, we keep making the model compact until it uses a similar amount of tokens compared to \method~ for a fair comparison. We use the same scaffold as ReasoningCache \citep{wu2026reasoningcachecontinualimprovement}, prompting the model to continue generating based on an existing trajectory or a summary of it. Detailed summarization and continuation prompts can be found in Appendix \ref{appendix:math_setup}. 

\begin{table}[h!]
\centering
\caption{Performance across competition math benchmarks, for fixed interval summary, we constrain the avg. number of tokens per question as the same as \method. Numbers in brackets indicate the average number of tokens used per question.}
\label{tab:performance_fixed_vs_adaptive}
\setlength{\tabcolsep}{4pt}
\begin{tabular}{lcccc}
\toprule
& \textbf{IMOBench} & \textbf{HMMT$_\text{Nov 25}$} & \textbf{HMMT$_\text{Feb 26}$} & \textbf{Avg.} \\
\midrule
\texttt{Qwen3-4B-Instruct-2507} \\[2pt]
\quad No Compaction [16k]          & 38.9$_{1.7}$ & 40.8$_{2.1}$ & 36.5$_{2.9}$ & 38.7$_{2.2}$ \\
\quad Fixed Interval Summary [44k] & 41.4$_{2.1}$ & 44.0$_{2.7}$ & 39.2$_{1.8}$ & 41.5$_{2.2}$ \\
\quad \method~[48k]                & \textbf{45.5}$_{2.8}$ & \textbf{47.8}$_{1.8}$ & \textbf{42.1}$_{2.8}$ & \textbf{45.1}$_{2.5}$ \\
\midrule
\texttt{Qwen3-30B-A3B-Instruct-2507} \\[2pt]
\quad No Compaction [16k]          & 45.2$_{1.2}$ & 54.7$_{2.7}$ & 51.9$_{2.8}$ & 50.6$_{2.2}$ \\
\quad Fixed Interval Summary [26k] & 48.7$_{2.9}$ & 57.3$_{1.7}$ & \textbf{58.7}$_{1.9}$ & 54.9$_{2.2}$ \\
\quad \method~[29k]                & \textbf{52.1}$_{2.6}$ & \textbf{59.4}$_{1.2}$ & 57.6$_{1.4}$ & \textbf{56.4}$_{1.7}$ \\
\midrule
\texttt{Qwen3.5-9B (Thinking Disabled)} \\[2pt]
\quad No Compaction [16k]          & 25.0$_{1.0}$ & 38.2$_{1.2}$ & 34.2$_{2.7}$ & 32.5$_{1.6}$ \\
\quad Fixed Interval Summary [90k] & 33.4$_{2.7}$ & 42.1$_{2.2}$ & 44.9$_{2.9}$ & 40.1$_{2.6}$ \\
\quad \method~[93k]                & \textbf{41.4}$_{2.3}$ & \textbf{48.2}$_{1.6}$ & \textbf{52.3}$_{2.0}$ & \textbf{47.3}$_{2.0}$ \\
\midrule
\texttt{Qwen3.5-4B (Thinking Disabled)} \\[2pt]
\quad No Compaction [16k]          & 16.9$_{3.0}$ & 26.4$_{1.2}$ & 22.5$_{1.6}$ & 21.9$_{1.9}$ \\
\quad Fixed Interval Summary [64k] & 27.4$_{0.9}$ & 35.5$_{2.0}$ & 29.1$_{2.9}$ & 30.7$_{1.9}$ \\
\quad \method~[67k]                & \textbf{34.1}$_{2.7}$ & \textbf{37.4}$_{1.1}$ & \textbf{30.0}$_{2.2}$ & \textbf{33.8}$_{2.0}$ \\
\bottomrule
\end{tabular}
\end{table}


\paragraph{Results.} Table \ref{tab:performance_fixed_vs_adaptive} shows the results across three competition math benchmarks and four model configurations. Under matched token budgets, \method~consistently outperforms both the baseline and fixed interval summarization, achieving the best performance in 11 out of 12 settings. The gains are most pronounced for the thinking-enabled models: on Qwen3.5-9B, \method~improves over the baseline by 16.4, 10.0, and 18.1 points on IMO-Answerbench, HMMT Nov, and HMMT Feb respectively. The one exception is \texttt{Qwen3-30B-A3B} on HMMT Feb, where fixed interval summarization edges out \method~by 1.1 points, though the latter still leads on the other two benchmarks. 


\begin{wraptable}[10]{r}{0.35\textwidth}
\vspace{-20pt}
\centering
\caption{Answer transitions across 12 fixed interval summarization calls for Qwen3-4B-Instruct-2507 on IMO-Answerbench.}
\vspace{10pt}
\label{tab:transitions}
\begin{tabular}{lc}
\toprule
Transition & Count \\
\midrule
Wrong $\rightarrow$ Correct & 1486 \\
Correct $\rightarrow$ Wrong & 1009 \\
\bottomrule
\end{tabular}
\vspace{-20pt}
\end{wraptable}
\paragraph{Headroom analysis.} We analyze the failure modes of fixed interval summarization for \texttt{Qwen3-4B-Instruct-2507} on IMO-Answerbench by tracking answer transitions across the 12 summarization calls. \autoref{tab:transitions} reports the number of instances where the model's answer changes from correct to wrong after a summarization and vice versa.

\begin{wraptable}[8]{r}{0.48\textwidth}
\vspace{-10pt}
\centering
\small
\caption{Oracle analysis on IMO-Answerbench for Qwen3-4B-Instruct-2507.}
\label{tab:oracle}
\begin{tabular}{lc}
\toprule
Method & Accuracy (\%) \\
\midrule
No Compaction [16k] & 38.9$_{1.7}$ \\
Fixed Interval Summary [44k] & 41.4$_{2.1}$ \\
\method~[48k] & 45.5$_{2.8}$ \\
Oracle (skip if correct) [44k] & 52.9$_{1.4}$ \\
\bottomrule
\end{tabular}
\end{wraptable}
While summarization produces a net gain, 40.4\% of all transitions cause degradations. This motivates a natural question: what if the model could selectively skip summarization when its current answer is already correct? \autoref{tab:oracle} shows that an oracle policy which suppresses summarization calls whenever the current answer is correct but otherwise follows the same fixed schedule achieves 52.9\%, a +11.5 improvement over fixed interval and +14.0 over baseline. Notably, this oracle is a strict subset of what a fully adaptive policy could achieve, as it only decides \emph{whether} to summarize at each fixed interval, not \emph{when} to summarize.

Both our results on Qwen3.5, where adaptive summarization already outperforms fixed interval, and this oracle analysis indicate substantial headroom for adaptive policies that learn when summarization is beneficial.

\subsection{Agentic Search}
\label{subsec:agentic-search}


\paragraph{Setup.}

We evaluate 3 models of different sizes and families: \texttt{GLM-4.7-Flash} (30B total, 3B active; \citep{5team2025glm45agenticreasoningcoding}), \texttt{MiniMax-M2.5} (230B total, 10B active; \citep{minimax2026m25}) and \texttt{Mimo-V2-Flash} (309B total, 15B active; \citep{coreteam2026mimov2flashtechnicalreport}) on 3 agentic search tasks: BrowseComp \citep{wei2025browsecompsimplechallengingbenchmark} , BrowseComp plus \citep{chen2025BrowseCompPlus} and DeepSearch QA \citep{gupta2026deepsearchqabridgingcomprehensivenessgap}. We use the exact same scaffold and evaluation settings as \citet{lee2026agenticaggregationparallelscaling}. We defer detailed sampling parameters and prompts to Appendix \ref{appendix:search_setup}.

For cost, we report the per-question USD cost. In the main text we use a
single-rate approximation that bills every prompt token at the cached
rate,
$\textsc{Cost}(q) = (p_{\text{cache}}\,N_{\text{prompt}} + p_{\text{out}}\,N_{\text{out}})/10^6$,
where $N_{\text{prompt}}$ and $N_{\text{out}}$ are the cumulative prompt
and completion tokens summed across every LLM call in the trajectory
(assistant turns, rubric probes, and summaries). This is conservative:
the first-time prefill tokens that a provider bills at the higher
uncached rate $p_{\text{in}}$ are only a small fraction (empirically
$5$--$8\%$) of $N_{\text{prompt}}$, so the effective rate stays close to
$p_{\text{cache}}$. Prices come from OpenRouter
(Table~\ref{tab:openrouter_pricing}). Appendix~\ref{appendix:cost} gives
the full procedure, and Appendix~\ref{appendix:tokens} reports the
finer-grained accounting that separates first-time prefill (billed at
$p_{\text{in}}$) from cached reads (billed at $p_{\text{cache}}$).
We compare \method~against the following baseline methods:

\begin{itemize}[leftmargin=*]
    \item \textbf{No Compaction}: No context management is used, model stops when either max context window is reached, or max number of tool calls are executed.
    \item  \textbf{Fixed-interval summary}: Summarization is triggered when 30\% of the max context window is consumed, a threshold suggested by \citet{liu2025webexplorer}, closely resembling \citet{wu2026reasoningcachecontinualimprovement}. 
    \item \textbf{Delete-all}: All the past trajectory is discarded when 30\% of the max context window is consumed.
    \item \textbf{Keep-last-N}: The last 3 turns are saved and the rest is discarded when 30\% of the max context window is consumed.
\end{itemize}

\paragraph{Main Results.} \autoref{tab:perf-cost-multi} shows the results of the three deployed agents (\texttt{GLM-4.7-Flash}, \texttt{MiniMax-M2.5}, \texttt{MiMo-V2-Flash}) on agentic search. \method~is the strongest method for all three models, improving over the no-compaction baseline by $+8.5$ (GLM-4.7-Flash), $+9.2$ (MiniMax-M2.5), and $+5.3$
  (Mimo-V2-Flash) absolute points on BrowseComp-Plus.

  The accuracy ordering is consistent across all three models: baseline $<$ fixed-interval $\le$ \method{}.
  Fixed-interval summarization recovers most of the baseline's lost ground; \method{}'s autonomous trigger adds up to $+6.3$ points by avoiding poorly-timed compressions the
   rubric concentrates that distribution at meaningful stopping points.

  On the cost axis, \method{} is \emph{cheaper} than the baseline despite issuing an extra LLM call at each probe. Per-question cost on BrowseComp-Plus drops by
  $67\%$ (GLM-4.7-Flash, $0.12 \to 0.04$), $63\%$ (MiniMax-M2.5, $0.19 \to 0.07$), and $33\%$ (Mimo-V2-Flash, $0.24 \to 0.16$). The KV-cache reuse described in \S3 is what
  makes the probe affordable: each rubric judgement contributes only its own short generation, since the trajectory's cache is preserved. The bulk of the savings come from the post-summarization regime, where every subsequent token attends to $\tilde{y}$ rather than $y_{1:t}$.
\begin{table}[h!]
\centering
\small
\caption{Performance and per-question cost across benchmarks. ``Cost''
is USD per question, computed with a single-rate approximation that
bills every prompt token at the cached rate,
$(p_{\text{cache}}\,N_{\text{prompt}} + p_{\text{out}}\,N_{\text{out}})/10^6$,
where $N_{\text{prompt}}$ and $N_{\text{out}}$ are the cumulative
prompt and completion tokens summed across every LLM call in the
trajectory; Appendix~\ref{appendix:tokens} reports the finer-grained
cost that separates first-time prefill from cached reads, and the full
bookkeeping is in Appendix~\ref{appendix:cost}.
We highlight the cost reduction \% compared to the ``no compaction''
baseline for \method{}. \textbf{Bold} indicates the best value per
column within each model block. Following
\citet{sun2025scalinglonghorizonllmagent} and
\citet{lee2026agenticaggregationparallelscaling}, evaluations are done
on 150 subsampled questions per benchmark.}
\label{tab:perf-cost-multi}
\setlength{\tabcolsep}{3pt}
\begin{NiceTabular}{llcccccccc}[cell-space-limits=2pt]
\CodeBefore
  \rectanglecolor{glmcolor}{3-1}{7-1}
  \rectanglecolor{minimaxcolor}{8-1}{12-1}
  \rectanglecolor{mimocolor}{13-1}{19-1}
\Body
\toprule
 & \multirow{3}{*}{\textbf{Context Management}} & \multicolumn{2}{c}{\textbf{BrowseComp}} & \multicolumn{2}{c}{\textbf{BrowseComp-Plus}} & \multicolumn{2}{c}{\textbf{DeepSearchQA}} & \multicolumn{2}{c}{\textbf{Overall}} \\
\cmidrule(lr){3-4}\cmidrule(lr){5-6}\cmidrule(lr){7-8}\cmidrule(lr){9-10}
 & & Acc. $\uparrow$ & Cost $\downarrow$ & Acc. $\uparrow$ & Cost $\downarrow$ & Acc. $\uparrow$ & Cost $\downarrow$ & Acc. $\uparrow$ & Cost $\downarrow$ \\
\midrule \midrule
\multirow{5}{*}{\rotatebox[origin=c]{90}{\footnotesize\texttt{ GLM-4.7-Flash }}}
& No Compaction               & 30.1$_{1.0}$ & 0.14 & 45.6$_{0.7}$          & 0.12          & 34.2$_{0.3}$ & 0.13 & 36.6$_{0.4}$ & 0.13 \\
& Fixed-interval & 35.4$_{1.8}$ & 0.06 & 50.0$_{2.3}$          & 0.04 & 39.1$_{0.3}$ & 0.05 & 41.5$_{1.0}$ & 0.05 \\
& Delete-all & 31.6$_{1.3}$ & \textbf{0.03} & 47.3$_{1.3}$ & \textbf{0.02} & 35.8$_{1.5}$ & \textbf{0.03} & 38.2$_{0.8}$ & \textbf{0.03} \\
& Keep-last-N & 33.5$_{2.4}$ & 0.05 & 51.4$_{1.6}$ & 0.03 & 38.8$_{0.5}$ & 0.04 & 41.2$_{1.0}$ & 0.04 \\
\cmidrule(lr){2-10}
& \method{} (Ours)    & \textbf{41.2}$_{2.3}$ & 0.10\,\textcolor{green!60!black}{\scriptsize(-29\%)} & \textbf{54.1}$_{1.4}$ & 0.04\,\textcolor{green!60!black}{\scriptsize(-67\%)} & \textbf{44.0}$_{1.3}$ & 0.07\,\textcolor{green!60!black}{\scriptsize(-46\%)} & \textbf{46.4}$_{1.0}$ & 0.07\,\textcolor{green!60!black}{\scriptsize(-46\%)} \\

\midrule
\multirow{5}{*}{\rotatebox[origin=c]{90}{\footnotesize\texttt{ MiniMax-M2.5 }}}
& No Compaction              & 47.2$_{0.7}$ & 0.19 & 62.0$_{1.8}$          & 0.19 & 52.0$_{1.4}$ & 0.19 & 54.6$_{0.8}$ & 0.19 \\
& Fixed-interval & 55.3$_{1.6}$ & 0.07 & 65.9$_{2.1}$          & 0.04 & 56.7$_{0.6}$ & 0.06 & 59.3$_{0.9}$ & 0.06 \\
& Delete-all & 52.7$_{1.3}$ & \textbf{0.05} & 65.0$_{1.3}$ & \textbf{0.03} & 54.9$_{0.7}$ & \textbf{0.04} & 57.5$_{0.7}$ & \textbf{0.04} \\
& Keep-last-N &  54.5$_{1.7}$ & 0.06 & 67.4$_{0.4}$ & 0.05 & 57.0$_{1.8}$ & 0.06 & 59.6$_{0.8}$ & 0.06 \\
\cmidrule(lr){2-10}
& \method{}  (Ours)   & \textbf{59.3}$_{1.4}$ & 0.09\,\textcolor{green!60!black}{\scriptsize(-53\%)} & \textbf{71.2}$_{2.6}$ & 0.07\,\textcolor{green!60!black}{\scriptsize(-63\%)} & \textbf{61.3}$_{2.2}$ & 0.08\,\textcolor{green!60!black}{\scriptsize(-58\%)} & \textbf{63.9}$_{1.2}$ & 0.08\,\textcolor{green!60!black}{\scriptsize(-58\%)} \\

\midrule
\multirow{6}{*}{\rotatebox[origin=c]{90}{\footnotesize\texttt{Mimo-V2-Flash}}}
& No Compaction               & 42.7$_{0.9}$ & 0.27 &    57.6$_{1.4}$         & 0.24   & 46.5$_{1.4}$ & 0.25 & 48.9$_{0.7}$ & 0.25 \\
& Fixed-interval & 51.6$_{2.2}$ & 0.14 &      60.2$_{1.9}$       & 0.14   & 52.3$_{1.4}$ & 0.14 & 54.7$_{1.1}$ & 0.14 \\
& Delete-all             & 45.5$_{1.1}$ & 0.08 & 61.2$_{2.3}$ & \textbf{0.11} & 49.7$_{2.1}$ & \textbf{0.09} & 52.1$_{1.1}$ & \textbf{0.09} \\
& Keep-last-N & 49.8$_{0.8}$ & \textbf{0.06} & \textbf{63.5}$_{1.7}$ & 0.15 & 53.0$_{0.5}$ & 0.10 & 55.4$_{0.6}$ & 0.10 \\
\cmidrule(lr){2-10}
& \method{} (Ours)     & \textbf{57.9}$_{2.4}$ & 0.11\,\textcolor{green!60!black}{\scriptsize(-59\%)} & 62.9$_{2.2}$ & 0.16\,\textcolor{green!60!black}{\scriptsize(-33\%)} & \textbf{56.8}$_{2.2}$ & 0.13\,\textcolor{green!60!black}{\scriptsize(-48\%)} & \textbf{59.2}$_{1.3}$ & 0.13\,\textcolor{green!60!black}{\scriptsize(-48\%)} \\

\bottomrule
\end{NiceTabular}
\vspace{-10pt}
\end{table}

\begin{figure}
    \centering
    \includegraphics[width=\linewidth]{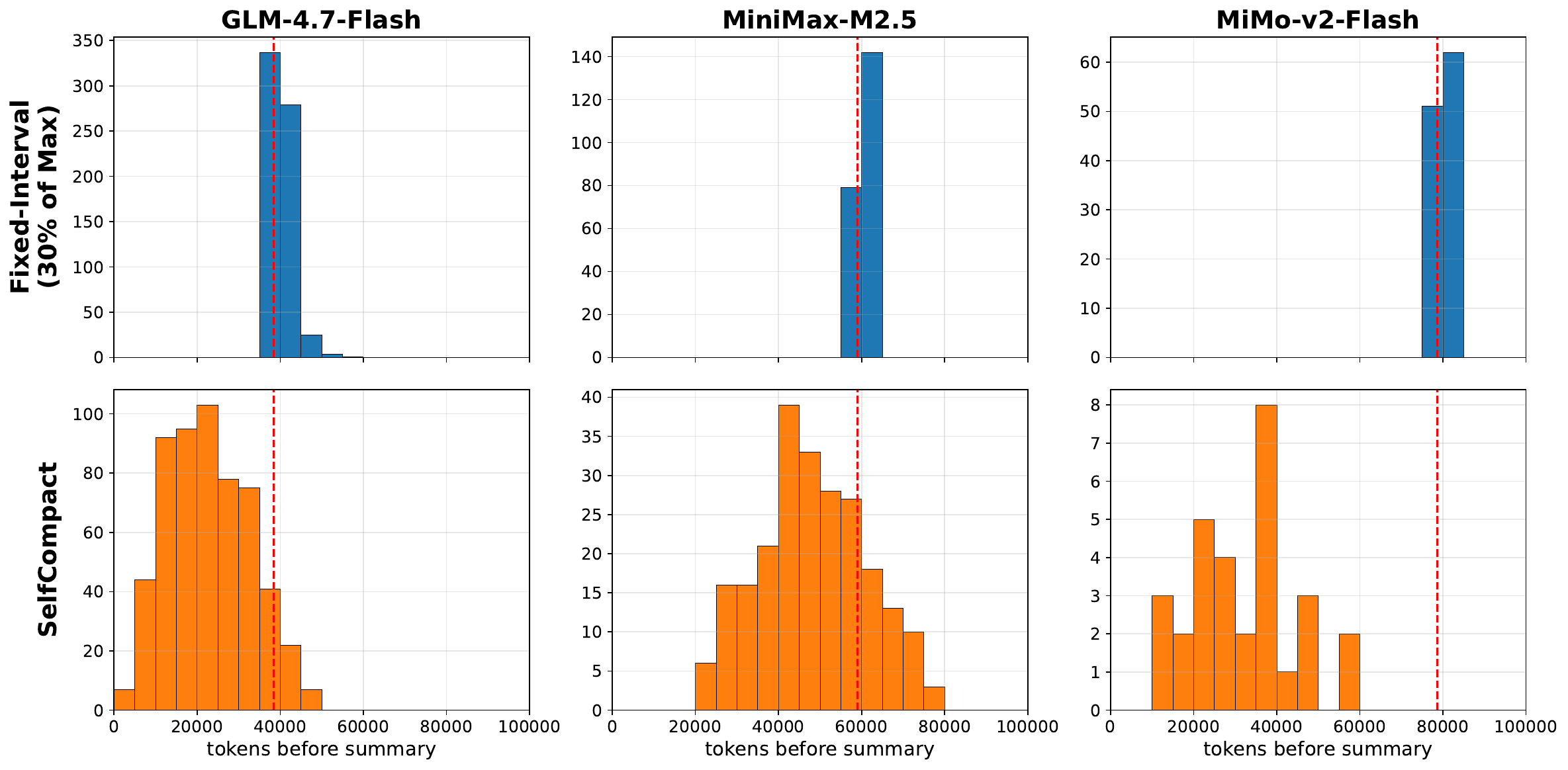}
    \caption{Distribution of context length when a summary fires in BrowseComp Plus. Top: Fixed-Interval: the policy compresses at a hard 30\%-of-max-context budget (red dashed). Bottom: \method{} --- the rubric fires when the model judges a sub-question resolved. Fixed-Interval triggers compaction at the threshold by construction; \method~spreads them across the budget.}
    \label{fig:tokens_before_summarization}
\end{figure}

\paragraph{\method{} fires compression earlier.}
  We plot the number of tokens accumulated before each compaction fires in Figure~\ref{fig:tokens_before_summarization}, comparing
  Fixed-Interval (30\% of context) against \method~ on BrowseComp-Plus. \method's rubric-fired summaries skew to the left of the 30\% threshold line in all three models, whereas Fixed-Interval pins
  triggers at the threshold by construction.
  This left skew indicates that the rubric reaches its ``compress now'' verdict well before the trajectory consumes 30\% of
  context. This indicates the fixed threshold typically fires too late, retaining stale tokens past the point where the model has already
  resolved the current sub-question.

\begin{figure}
    \centering
    \includegraphics[width=\linewidth]{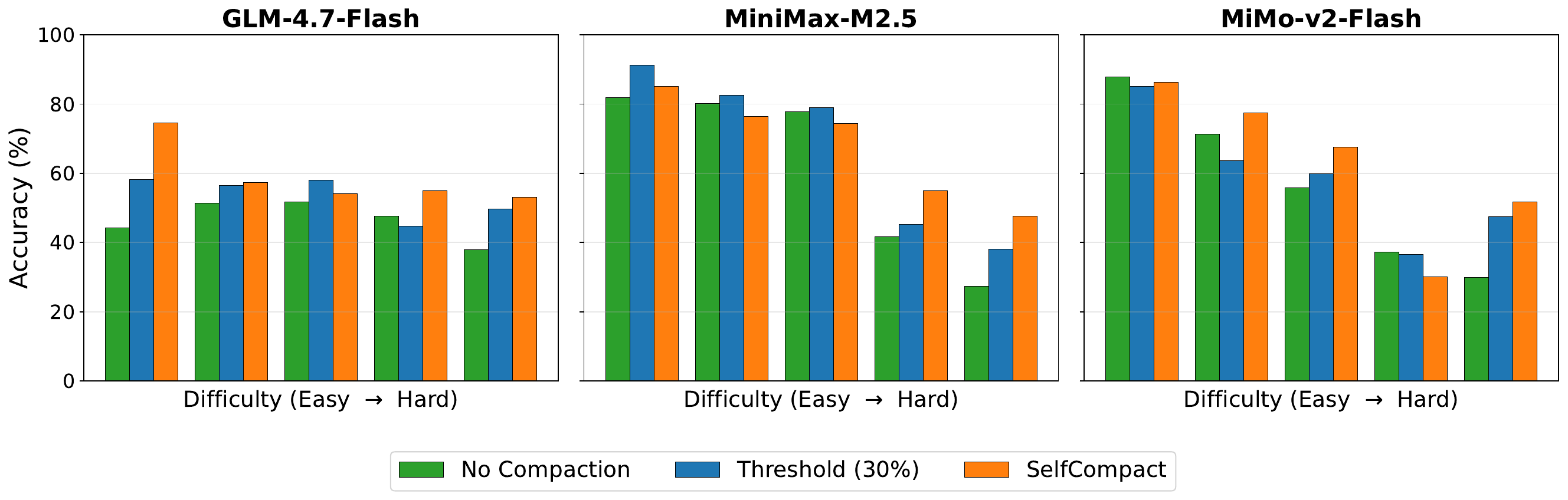}
    \caption{Accuracy by per-question difficulty on BrowseComp Plus for three models. Difficulty is the total output tokens consumed by the no-compression baseline, split into 5 equal-size quantile bins per model (left = easy, right = hard). Bars compare Baseline (no compression), a fixed-budget Threshold (30\% of context), and \method. \method~matches the others on easy bins and pulls ahead on the hardest bins across all three models.}
    \label{fig:difficulty_binned}
\end{figure}

\paragraph{\method{} helps on harder problems.}   We bin each question by the total output tokens its no-compression baseline
  trajectory consumed (a model-specific proxy for question difficulty) and
  split into five equal-size quantiles per model. Within each bin we report
  accuracy for the baseline, a fixed-budget Threshold (30\% of context), and
  \method{} (Figure~\ref{fig:difficulty_binned}). On easy bins the three
  policies are within sampling noise, but on the two hardest bins
  \method{} improves over the Threshold baseline by 5--20\,pp consistently
  across 3 models, indicating that
  rubric-driven compression helps the most on questions that require deep
  search and accumulate the largest contexts.

\paragraph{Ablation on the effect of rubrics.} We ablate the effect of rubrics by removing the rubric-gated verification and instead letting the model self-decide whether to summarize at each step. This collapses performance to 41.0\% on average with \texttt{GLM-4.7-Flash} (\autoref{tab:selfcompact-glm}), a 5.4-point drop from full SelfCompact (46.4\%) that leaves it on par with the naive fixed-interval schedule (41.5\%) and below it on BrowseComp (33.6 vs 35.4), with the same effect on math reasoning using \texttt{Qwen3-4B-Instruct-2507} (40.9\% vs 45.5\%, \autoref{tab:selfcompact-math}). This shows that the gains do not come from triggering compaction per se, but from the rubric-gated verification that determines what reasoning state is safe to preserve; without it, the model's unconstrained summarization discards information that later steps depend on. In other words, knowing when to compact is insufficient, and the rubrics are what make the retained context reliable.

\begin{table}[t]
\centering
\caption{Accuracy (\%) across agentic search benchmarks with \texttt{GLM-4.7-Flash}. Subscripts denote standard deviation over 5 runs. Best per column in bold.}
\label{tab:selfcompact-glm}
\begin{tabular}{lcccc}
\toprule
Method & \textbf{BrowseComp} & \textbf{BrowseComp-Plus} & \textbf{DeepSearchQA} & \textbf{Avg.} \\
\midrule
No Compaction           & 30.1$_{1.0}$ & 45.6$_{0.7}$ & 34.2$_{0.3}$ & 36.6$_{0.4}$ \\
Fixed-interval          & 35.4$_{1.8}$ & 50.0$_{2.3}$ & 39.1$_{0.3}$ & 41.5$_{1.0}$ \\
\method~w/o rubrics & 33.6$_{2.1}$ & 51.9$_{2.4}$ & 37.6$_{1.9}$ & 41.0$_{1.2}$ \\ \midrule
\method          & \textbf{41.2}$_{2.3}$ & \textbf{54.1}$_{1.4}$ & \textbf{44.0}$_{1.3}$ & \textbf{46.4}$_{1.0}$ \\
\bottomrule
\end{tabular}
\end{table}

\section{Related Work}


\paragraph{Context compaction in frontier models.} In API-based models, context compaction has become a standard built-in feature to prevent multi-rounds of long reasoning chains from blowing up the context window.\footnote{e.g. \url{https://platform.claude.com/cookbook/tool-use-automatic-context-compaction}}
Note that while the feature is often named as ``auto compaction'', it refers to the compaction operation being automatically triggered and performed when the \emph{context length reaches a target threshold}, i.e.\ fixed interval.
Such mechanisms have raised complaints from the community\footnote{\url{https://www.reddit.com/r/codex/comments/1qib69i/auto_compaction_is_not_that_helpful/}}, where compaction at inappropriate positions tends to remove important information from the context, which is a central issue we aim to resolve.
The only exception is a new feature from LangChain posted concurrently to our work\footnote{\url{https://www.langchain.com/blog/autonomous-context-compression}} March 2026, where they also introduced a summary tool; they did not introduce the rubric-based criteria for deciding \textit{when} to summarize.

\begin{wraptable}{r}{0.42\textwidth}
\vspace{-20pt}
\centering
\caption{Accuracy (\%) on IMOBench with \texttt{Qwen3-4B-Instruct-2507.} Subscripts denote standard deviation.}
\vspace{5pt}
\label{tab:selfcompact-math}
\begin{tabular}{lc}
\toprule
Method & Math \\
\midrule
No Compaction                & 38.9$_{1.7}$ \\
Fixed-interval          & 41.4$_{2.1}$ \\
\method~w/o rubrics & 40.9$_{2.3}$ \\ \midrule
\method            & \textbf{45.5}$_{2.8}$ \\
\bottomrule
\end{tabular}
\vspace{2pt}
\end{wraptable}
\paragraph{Learning to compact during post-training.}
Long-horizon reasoning and agentic search accumulate tokens at an unsustainable rate: observations pile up, intermediate results go stale, and prior context anchors rather than informs subsequent generations. A broad line of work addresses this through periodic compaction, sharing a common paradigm: generate an extended sequence of thoughts or actions, summarize the accumulated context, resume from the compressed state, and repeat \citep{shao2025foldactefficientstablecontext,sun2025scalinglonghorizonllmagent}. The majority of these methods \textit{train} models to perform compaction (via SFT, RL, or both) in order to  encode the when-to-compact decision implicitly in teacher demonstrations or rewards~\citep{kontonis2026memento,yang2026accordionthinkingselfregulatedstepsummaries,yan2026inftythink, 
cassano2026selfsummarization,mitra2025motif,aghajohari2025markovian, wu2026reasoningcachecontinualimprovement, sun2025scalinglonghorizonllmagent, ye2025agentfoldlonghorizonwebagents, shao2025foldactefficientstablecontext, wu2026resumunlockinglonghorizonsearch}. 
Our approach is entirely training-free. We draw on recent advances in tool-using models~\citep{5team2025glm45agenticreasoningcoding,minimax2026m25}, which have unlocked the ability to invoke external tools mid-generation, and show that off-the-shelf models can self-compact effectively if scaffolded correctly. A lightweight rubric specifying when to invoke the compaction tool is sufficient to elicit reliable adaptive compaction without any parameter updates.

Across these methods, most also trigger compaction at \textit{fixed} token-count thresholds independent of trajectory structure \citep{yan2026inftythink, aghajohari2025markovian, wu2026reasoningcachecontinualimprovement, wu2026resumunlockinglonghorizonsearch}, risking compaction mid-derivation when the model still needs the context it is about to discard. Our rubric is designed precisely to detect these structural conditions, firing when a sub-task has resolved or the trajectory is converging, and suppressing when the model is mid-derivation or stuck.

Few works have studied compaction without model training. The closest to ours is by  \citet{zhu2026retracrecursivetrajectorycompression}, which introduces a training-free  variant that uses a frontier model as an external summarizer applied at \textit{fixed} intervals after each full search trajectory. Our work differs in that, we introduce a rubric that dynamically determines when compaction should fire rather than relying on fixed post-trajectory intervals. Our work also connects to a broader goal of context management. We refer the interested reader to the existing surveys on this topic~\citep{mei2025survey}. 

\paragraph{KV Cache Eviction.}
A complementary line of work reduces memory and compute costs by evicting or compressing KV cache entries during inference. 
Static methods retain tokens based on recency \citep{xiao2023efficient} or attention scores \citep{li2024snapkv}, while more recent approaches use richer importance measures that are often learned \citep{park2025keydiff,eyuboglu2025cartridges, zweiger2026fast,monea2025breadcrumbs,li2026neuralgarbagecollectionlearning}. Across this literature, interpretability of the compressed representation is treated as secondary: what is retained or evicted need not be human-readable, only attention-efficient, and may even be complemented by black-box weight updates \citep{zuo2026rethinkingloramemorylens}. The context compaction literature (which is where our work fits in), by contrast, prioritizes discrete natural-language summaries that remain inspectable and overridden.

\section{Limitations}
\label{sec:limitations}

\paragraph{Model capability gap.} We only evaluated open-weight models;
frontier systems (GPT-5.5~\citep{openai2026gpt55}, Claude Opus
4.7~\citep{anthropic2026opus47}, Gemini~3~Pro~\citep{google2026gemini31pro})
report stronger metacognition~\citep{lindsey2025introspection} and may
detect context rot without a rubric. \method{} is complementary: a
training-free scaffold that layers onto any model and closes the gap
precisely where it exists in deployed open-weight agents.


\paragraph{No reinforcement learning.} We scope this work to
training-free interventions to isolate the rubric's contribution. Prior
work~\citep{kontonis2026memento, zhu2026lightthinkerreasoningcompressionmemory} shows RL can teach a model both
\emph{when} and \emph{what} to compact. We view RL as a natural
extension: the rubric supplies a behavioral target that RL can distill
into the policy.

\section{Conclusion}

We introduced \method{}, a rubric-gated summarization scaffold that
compacts agent trajectories at closed reasoning units rather than on
fixed intervals. Across six benchmarks and seven models, it matches
or exceeds fixed-interval summarization at 30--70\% lower cost,
reframing \emph{when to compact} as a capability that a lightweight
rubric can supply without training.

\section*{Acknowledgments}
This work was supported in part by Apple and Defense Advance Research Projects Agency
(DARPA) under Contract No.  HR001125C0304
and ONR grant (N0001424-1-2089). 
Any opinions,
findings and conclusions or recommendations expressed in this material are those of the author(s)
and do not necessarily reflect the views of DARPA.
We acknowledge the use of computational resources from the Johns Hopkins DSAI cluster.

\bibliographystyle{plainnat}
\bibliography{refs}


\clearpage
\appendix

\section{Math experimental setup}
\label{appendix:math_setup}

\paragraph{Models.} We evaluate four open-weight Qwen models on competition
math: two instruction-tuned models (\texttt{Qwen3-4B-Instruct-2507} and
\texttt{Qwen3-30B-A3B-Instruct-2507}) and two thinking-disabled variants
(\texttt{Qwen3.5-4B} and \texttt{Qwen3.5-9B}). All checkpoints are loaded
from Hugging Face and served via vLLM~\citep{kwon2023efficient}. The
30B-A3B model uses tensor-parallel size 2; all others run on a single H200
GPU.

\paragraph{Benchmarks.} IMO-Answerbench~\citep{luong-etal-2025-towards},
HMMT November 2025, and HMMT February 2026. Final answers are extracted
and scored with the
\href{https://github.com/huggingface/Math-Verify}{\texttt{math\_verify}}
package.

\paragraph{Sampling parameters.} We use temperature $1.0$ and top-$p$
$0.7$ for all models, generating $n=16$ samples per question and
reporting mean$\,\pm\,$std over samples. Each generation round is capped
at $16{,}384$ tokens (\texttt{--num\_tokens 16384}); the per-trajectory
round budget is \texttt{--max\_rounds 12}, and the summarizer output is
hard-truncated at $512$ tokens (\texttt{--max\_summary\_tokens 512}).

\paragraph{Scaffold.} We adopt the ReasoningCache scaffold of
\citet{wu2026reasoningcachecontinualimprovement}: at each round, the
model is prompted to either solve from scratch or improve on a summary
of the previous attempt. The continuation prompt issued to the model
after every refinement is reproduced verbatim in Box~\ref{box:math_continuation}.

\paragraph{Methods.}
\textsc{Baseline} runs the model with a single $16$k-token budget
(\texttt{--max\_rounds 1}) and no summarization.
\textsc{Fixed-Interval Summary} sets \texttt{--force\_tool\_calls},
which fires the summarizer at the end of every round (i.e.\ every
$16{,}384$ generated tokens) regardless of trajectory state.
\textsc{\method{}} exposes \texttt{<tool\_call>} as a vLLM stop token so
that the model autonomously emits the summarizer; the round ends and the
summary fires whenever the stop token appears.
\textsc{\method{} w/ Rubrics} interposes the rubric probe in
Box~\ref{box:rubric_math} between rounds: if the rubric fires
\textsc{compress} the summarizer is invoked, otherwise the model
continues from the unmodified prefix.

  \begin{tcolorbox}[
  colback=teal!5,
  colframe=teal!70!black,
  coltitle=white,
  title={\textbf{Math rubric prompt} (\textsc{ANSWER} / \textsc{STUCK} / \textsc{HAS-NEXT}; appended to $(x, y_{1:t})$ at every round boundary)},
  label=box:rubric_math,
  fonttitle=\small,
  boxrule=0.5pt,
  arc=2pt,        
  left=6pt, right=6pt  
  ]
  \small
Judge your math-solving state from the conversation above. Answer Q1\dots Q3 with Y or N and one short evidence phrase. Answers without evidence must be N. ``Non-trivial fact'' = a new equation, reduction, bound, case-elimination, or counterexample, not a paraphrase.

  \medskip
  \textbf{Q1 ANSWER:} The latest round states a specific final answer (a \verb|\boxed{}| expression or ``Final Answer: \dots''), not merely a partial result.
    \emph{If Y}, quote the answer verbatim.
    \emph{If N}, state what is still unknown.\\

  \textbf{Q2 STUCK:} Your last 2 rounds added no non-trivial fact --- only paraphrases or abandoned attempts.
    \emph{If Y}, name the two rounds and write ``no new fact''.
    \emph{If N}, name one non-trivial fact from the last 2 rounds.\\

  \textbf{Q3 HAS-NEXT:} You can write the exact next step (case split, substitution, verification, lemma to prove).
    \emph{If Y}, write the step as one imperative sentence.
    \emph{If N}, write ``NONE''.\\

  \smallskip
  Output: exactly 3 lines, no preamble or trailing text.
  \begin{verbatim}
  Q1: Y/N -- <evidence>
  Q2: Y/N -- <evidence>
  Q3: Y/N -- <evidence>
  \end{verbatim}

  Fire rule: \textsc{compress} iff $\text{Q1}=\text{Y} \lor (\text{Q2}=\text{Y} \land \text{Q3}=\text{Y})$. Branch~A ($\text{Q1}=\text{Y}$) fires a lock-in re-prompt that preserves the boxed answer; Branch~B ($\text{Q2}=\text{Y} \land \text{Q3}=\text{Y}$) fires a reset that preserves the named next step.
  \end{tcolorbox}

  \begin{tcolorbox}[
  colback=teal!5,
  colframe=teal!70!black,
  coltitle=white,
  title={\textbf{Math summarizer prompt} (preserve-answer v3; replaces $y_{1:t}$ with $\tilde y$ when fired)},
  label=box:math_summarizer,
  fonttitle=\small,
  boxrule=0.5pt,
  arc=2pt,        
  left=6pt, right=6pt  
  ]
  \small
\textbf{Context Refinement Prompt:}

  \medskip
  \textbf{Original Prompt:} \texttt{\{original\_prompt\}}\\
  \textbf{Partial Generation:} \texttt{\{partial\_generation\}}

  \medskip
Your task is to create a compressed summary for another model to continue solving from.

  \medskip
  \textbf{RULES:}
  \begin{enumerate}\setlength\itemsep{0pt}\setlength\topsep{0pt}
  \item If a final answer (e.g., \verb|\boxed{}|) was found, PRESERVE IT at the end of your summary.
  \item Keep key insights, important calculations, and the reasoning path.
  \item Remove redundant text, false starts, and unnecessary repetition.
  \item If the answer seems wrong or unverified, note that verification is needed.
  \item Be concise but preserve all critical mathematical steps.
  \end{enumerate}

  \textbf{Output format:}
  \begin{itemize}\setlength\itemsep{0pt}\setlength\topsep{0pt}
  \item Key insights and progress made.
  \item Important intermediate results.
  \item If found: ``Final Answer: [the answer]'' or the \verb|\boxed{}| expression.
  \item If not solved: what still needs to be done.
  \end{itemize}
  \end{tcolorbox}

  \begin{tcolorbox}[
  colback=teal!5,
  colframe=teal!70!black,
  coltitle=white,
  title={\textbf{Continuation prompt} (issued to $\pi$ at the start of every round, with $\tilde y$ filled in if available)},
  label=box:math_continuation,
  fonttitle=\small,
  boxrule=0.5pt,
  arc=2pt,        
  left=6pt, right=6pt  
  ]
  \small
You are given a maths problem. You may also be given a summary of a previous attempt to solve it. This previous attempt may or may not be correct.

  \medskip
  \textbf{\#\#\# PROBLEM} \texttt{\{problem\}}\\
  \textbf{\#\#\# SUMMARY OF PREVIOUS ATTEMPT} \texttt{\{summary\}}

  \medskip
  \textbf{\#\#\# INSTRUCTIONS}

If no summary of a previous attempt is provided, solve the problem from scratch.

If a summary of a previous attempt is provided, your task is to improve upon this attempt. You should rely on this summary to guide your thinking. Some strategies you could use include:
  \begin{itemize}\setlength\itemsep{0pt}\setlength\topsep{0pt}
  \item Verifying the previous solution.
  \item Proving the result in a different way.
  \item Finding alternative problem-solving strategies.
  \item Continuing from where the previous solution left off, assuming that the previous solution is incomplete.
  \end{itemize}

Reason step-by-step and return your final answer in \verb|\boxed{}|.
  \end{tcolorbox}


\clearpage

\section{Agentic-search experimental setup}
\label{appendix:search_setup}

\paragraph{Models and serving.} The three deployed agents
(\texttt{GLM-4.7-Flash}, \texttt{MiniMax-M2.5}, \texttt{Mimo-V2-Flash})
are accessed through OpenRouter via the AggAgent
runtime~\citep{lee2026agenticaggregationparallelscaling}. We use the
unmodified
\texttt{react\_agent\_selfcheck} scaffold; the rubric and summarizer
prompts are reproduced verbatim in Box~\ref{box:rubric} and
Box~\ref{box:summ}.

\paragraph{Benchmarks.} BrowseComp~\citep{wei2025browsecompsimplechallengingbenchmark},
BrowseComp-Plus~\citep{chen2025BrowseCompPlus}, and
DeepSearchQA~\citep{gupta2026deepsearchqabridgingcomprehensivenessgap}.
Following \citet{sun2025scalinglonghorizonllmagent} and
\citet{lee2026agenticaggregationparallelscaling}, we sub-sample 150
questions per benchmark.

\paragraph{Sampling parameters.} All three models share the same
decoding configuration: temperature $1.0$, top-$p$ $0.95$, max output
tokens per call $10{,}000$, parallel-tool-calls disabled, and a
per-trajectory cap of $100$ LLM calls. Per-model context windows, the
$30\%$-of-context trigger used by \textsc{Fixed-Interval Summary} and
the rubric's token-pct backstop, and model-specific decoding flags are
listed in Table~\ref{tab:search_model_config}.

\begin{table}[h!]
\centering
\small
\caption{Per-model context windows, $30\%$-of-context summarization
trigger, and decoding flags for the three OpenRouter agents.
GLM-4.7-Flash keeps the thinking trace visible to the rubric probe via
\texttt{enable\_thinking=True} and \texttt{clear\_thinking=False}.}
\label{tab:search_model_config}
\setlength{\tabcolsep}{6pt}
\begin{tabular}{lrrl}
\toprule
\textbf{Model} & \textbf{Ctx window} & \textbf{$30\%$ trigger}
& \textbf{Model-specific flags} \\
\midrule
\texttt{GLM-4.7-Flash}  & $128{,}000$ & $38{,}400$
  & \texttt{enable\_thinking=True}, \texttt{clear\_thinking=False} \\
\texttt{MiniMax-M2.5}   & $196{,}608$ & $58{,}982$
  & \texttt{reasoning\_split=True} \\
\texttt{Mimo-V2-Flash}  & $262{,}144$ & $78{,}643$
  & (default OpenRouter body) \\
\bottomrule
\end{tabular}
\end{table}

\paragraph{Scaffold details.} A single trajectory is the standard ReAct
loop: the agent emits a tool call (\texttt{search} and either
\texttt{visit} for BrowseComp/DeepSearchQA or
\texttt{get\_document\_bcp} for BrowseComp-Plus), receives the result,
reasons, and either issues another tool call or commits a final answer.
For BrowseComp and DeepSearchQA, \texttt{search} is backed by the
Serper API (\url{https://serper.dev/}) and \texttt{visit} fetches and
extracts page content via crawl4ai
(\url{https://github.com/unclecode/crawl4ai}); for BrowseComp-Plus,
\texttt{search} and \texttt{get\_document\_bcp} resolve against the
benchmark's released corpus.
The rubric in Box~\ref{box:rubric} is appended to a \emph{copy} of the
trajectory at each round boundary so the probe does not pollute the
rolling cache. \textsc{compress} fires only when four gates pass
simultaneously: (\textsc{round}) iteration $\geq 3$,
(\textsc{tokens}) running prompt length $\geq 40{,}000$,
(\textsc{cap}) total summaries so far $< 1$, and (\textsc{period})
$\geq 2$ rounds elapsed since the last probe. The token-pct backstop
forces a \textsc{compress} once the prompt crosses $0.30 \cdot
\text{ctx\_window}$ regardless of the rubric.

  \begin{tcolorbox}[
  colback=teal!5,
  colframe=teal!70!black,
  coltitle=white,
  title={\textbf{Search rubric probe} (appended to $(x, y_{1:t})$ at every round boundary)},
  label=box:rubric,
  fonttitle=\small,
  boxrule=0.5pt,
  arc=2pt,        
  left=6pt, right=6pt  
  ]
  \small
      You are about to decide whether to compress your conversation history into a summary that REPLACES the full history above. After compression, you continue research from
  only \texttt{[system, original\_question, assistant\_summary, user\_continue]}. Compression is irreversible: anything not preserved in the summary is gone. \\

  \smallskip
  Compression is safe ONLY when ALL FOUR of the following hold:
  \begin{itemize}\setlength\itemsep{0pt}\setlength\topsep{0pt}
  \item[(C1)] the trajectory has reached a closed unit (not mid-thought),
  \item[(C2)] the essential information is reducible to 3--5 cite-able facts without loss,
  \item[(C3)] something has progressed since the last compression,
  \item[(N1)] you are NOT currently stuck in a way summarization would mask.
  \end{itemize}

  Answer C1, C2, C3, N1 honestly. Each Y answer requires verbatim evidence quoted from the trajectory above; answers without evidence default to N. \\

  \smallskip
  \textbf{C1 CLOSED-UNIT:} The most recent assistant message is a closed unit --- a completed tool call whose result is now visible, or a completed sub-analysis with a clear
   stopping point. It is NOT mid-sentence reasoning (``Let me now check\dots'', ``I should next look at\dots''), and not a half-formulated query.
    \emph{If Y}, quote the closing fragment of the last assistant message.
    \emph{If N}, quote the open fragment that shows the trajectory is mid-thought.\\

  \textbf{C2 SUMMARIZABLE:} You can write 3--5 essential facts (with verbatim citations from the trajectory) that future-you needs to continue research after compression.
  Each fact must be a single concrete statement: a name, date, URL, quoted claim, or resolved sub-question. Answer N if the trajectory's value is dispersed across many small
   inferences (e.g., a list of dead-end queries needed to avoid retries, negative results that constrain hypothesis space) that would be lost without the dispersal.
    \emph{If Y}, list the 3--5 facts numbered, each with a verbatim citation in quotes, separated by ``$\,|\,$''.
    \emph{If N}, name in one sentence the class of information that would be lost.\\

  \textbf{C3 PROGRESS:} Since the most recent compression (or since the start of the conversation if none), you have either obtained a new concrete fact (name, date, URL, or
   quoted claim) OR refined the sub-question being pursued.
    \emph{If Y}, name the new fact or refined sub-question.
    \emph{If N}, state that you are returning the same state you compressed from.\\

  \textbf{N1 STUCK:} At least 3 of your last 4 search queries returned no new URL or fact (i.e., were duplicates or returned already-known content). If you have made fewer
  than 4 searches total, answer N.
    \emph{If Y}, name 1 distinct strategy you have NOT yet tried (different tool, different query type, different angle on the question).
    \emph{If N}, name one new URL or fact obtained recently. \\

  \smallskip
  Output: exactly 4 lines, no preamble or trailing text.
  \begin{verbatim}
  C1: Y/N -- <evidence>
  C2: Y/N -- <if Y: 1. fact "citation" | 2. fact "citation" |
              3. fact "citation"; if N: <class of info lost>>
  C3: Y/N -- <evidence>
  N1: Y/N -- <evidence>
  \end{verbatim}

  Fire rule: \textsc{compress} iff $\text{C1}=\text{Y} \land \text{C2}=\text{Y} \land \text{C3}=\text{Y} \land \text{N1}=\text{N}$; otherwise \textsc{continue}.
  \end{tcolorbox}

\paragraph{Summarizer.} The search summarizer is the
\texttt{webresummer} variant of
\citet{wu2026resumunlockinglonghorizonsearch}, reproduced verbatim in
Box~\ref{box:summ} below; we use it unmodified for all three deployed agents.

  \begin{tcolorbox}[
  colback=teal!5,
  colframe=teal!70!black,
  coltitle=white,
  title={\textbf{Summarizer prompt} ($\mathcal{S}$ instruction appended to $(x, y_{1:t})$)},
  label=box:summ,
  fonttitle=\small,
  boxrule=0.5pt,
  arc=2pt,        
  left=6pt, right=6pt  
  ]
  \small
You are an expert at analyzing conversation history and extracting relevant information. Your task is to thoroughly evaluate the conversation history above and the user's original question to provide a summary that will REPLACE the full conversation history when you continue.

  \medskip
  \textbf{Task Guidelines}

  \textbf{1. Information Analysis:}
  \begin{itemize}\setlength\itemsep{0pt}\setlength\topsep{0pt}
  \item Carefully analyze the conversation history to identify truly useful information.
  \item Focus on information that directly contributes to answering the question.
  \item Do NOT make assumptions, guesses, or inferences beyond what is explicitly stated in the conversation.
  \item If information is missing or unclear, do NOT include it in your summary.
  \end{itemize}

  \textbf{2. Summary Requirements:}
  \begin{itemize}\setlength\itemsep{0pt}\setlength\topsep{0pt}
  \item Extract only the most relevant information that is explicitly present in the conversation.
  \item Synthesize information from multiple exchanges when relevant.
  \item Only include information that is certain and clearly stated in the conversation.
  \item Do NOT output or mention any information that is uncertain, insufficient, or cannot be confirmed from the conversation.
  \end{itemize}
  \end{tcolorbox}


\clearpage

\section{Cost analysis of summarization}
\label{appendix:cost}

\paragraph{Cost model (single-rate).} In the main text, for all
OpenRouter-hosted agents we charge each question with the additive
token-cost
\[
\textsc{Cost}(q)
= \frac{1}{10^6}\Big(
  p_{\text{cache}} \cdot N_{\text{prompt}}
  + p_{\text{out}} \cdot N_{\text{out}}
\Big),
\]
where $N_{\text{prompt}}$ is the cumulative prompt tokens summed across \emph{every} LLM call in
the trajectory assistant turns, the rubric probe, and the
summarizer call, and $N_{\text{out}}$ is the cumulative completion
tokens over the same set of calls.
We charge every prompt token at the cache rate $p_{\text{cache}}$
rather than separately tracking which tokens were a cache hit and which
were a first-time prefill. This is a conservative single-rate
approximation: the first-time prefill tokens --- the question, each
assistant turn, and each tool observation, counted once when they first
enter the prefix --- are only a small fraction (empirically $5$--$8\%$)
of $N_{\text{prompt}}$, which counts each token once per call it
survives, so the effective per-token rate sits close to
$p_{\text{cache}}$ even when provider-side caching misses occasionally.
Appendix~\ref{appendix:tokens} relaxes this approximation, splitting
$N_{\text{prompt}}$ into first-time prefill $N_{\text{prefill}}$ (billed
at the uncached rate $p_{\text{in}}$) and cache reads $N_{\text{cached}}$
(billed at $p_{\text{cache}}$) and reporting the resulting two-rate cost
in Table~\ref{tab:tokens_all}. Per-1M token prices are
listed in Table~\ref{tab:openrouter_pricing}.

\paragraph{Calculation procedure.} For each of the $150$ sampled
questions per benchmark, we (i)~load the trajectory's per-call usage
records logged by the AggAgent runtime, (ii)~accumulate
$(N_{\text{prompt}}, N_{\text{out}})$ as defined above,
(iii)~apply the model's prices from
Table~\ref{tab:openrouter_pricing}, and (iv)~report the mean across
the $150$ questions in Table~\ref{tab:perf-cost-multi} (units: USD per
question).

\begin{table}[h!]
\centering
\caption{OpenRouter prices in USD per $1{,}000{,}000$ tokens.}
\label{tab:openrouter_pricing}
\setlength{\tabcolsep}{8pt}
\begin{tabular}{lccc}
\toprule
\textbf{Model} & $p_{\text{in}}$ & $p_{\text{cache}}$ & $p_{\text{out}}$ \\
\midrule
\texttt{GLM-4.7-Flash}   & 0.07 & 0.01 & 0.40 \\
\texttt{MiMo-V2-Flash}   & 0.10 & 0.01 & 0.30 \\
\texttt{MiniMax-M2.5}    & 0.30 & 0.03 & 1.20 \\
\bottomrule
\end{tabular}
\end{table}

\paragraph{Where compaction saves cost.} Under the single-rate cache
billing above, the savings from compaction come entirely from
shortening every \emph{subsequent} call's prompt. After a fire that
collapses a $T$-token prefix to a $\tilde t$-token summary, every later
call pays $p_{\text{cache}} \cdot \tilde t$ instead of
$p_{\text{cache}} \cdot T$, a $T/\tilde t$ ratio per call.
Empirically summarization collapses a $50$--$100$k trajectory into
$\tilde t \approx 1$--$3$k, giving $20$--$80\times$ shrinkage of the
post-compact prompt. The rubric probe and summarizer both append to
the live prefix so the running prompt is preserved during the probe
itself; the probe contributes only its own $\sim$60-token verdict to
$N_{\text{out}}$, and the summarizer contributes $\tilde t$ tokens
once, both negligible relative to the future-call savings amortized
across the rest of the trajectory.

\paragraph{Math.} The Qwen runs are served locally with vLLM, so cost
collapses to a wall-clock trade-off rather than a dollar one. We report
the per-question budget tag \texttt{[Xk]} in
Table~\ref{tab:performance_fixed_vs_adaptive}: the average sum of
generated tokens (\texttt{rounds} $\times$ \texttt{num\_tokens}) plus
refinement tokens across the $n=16$ samples. \textsc{Fixed-Interval
Summary}'s budget is matched to \method{} within $\pm 3$k tokens so
all four rows in each model block compete at identical compute.


\clearpage

\section{Trajectory token consumption}
\label{appendix:tokens}

\paragraph{Token lifecycle.} A trajectory under \method{} looks like
$Q \to A_1 \to A_2 \to \cdots \to A_k \to S \to A'_1 \to A'_2 \to \cdots$,
where $Q$ is the original question (plus system prompt), each $A_t$
is one assistant turn (model output for tool call or final answer),
and $S$ is a summary the model emits when the rubric fires
\textsc{compress}. Each token a model produces is billed three different
ways over its lifetime:
\begin{itemize}[leftmargin=*]\setlength\itemsep{0pt}
\item \emph{Output} — when the model first emits the tokens (one charge).
\item \emph{Prefill} (uncached input) — on the immediately next LLM call,
when the tokens are appended to the prefix and seen by the provider for
the first time (one charge).
\item \emph{Cached input} — on every subsequent LLM call until the
prefix is reset (one charge per reuse).
\end{itemize}
Concretely, a single $A_t$ at round $t$ in a trajectory of $R$ rounds
contributes once to \emph{Output}, once to \emph{Prefill}, and $R-t$
times to \emph{Cached}. A summary $S$ emitted at compaction is billed
once as output, once as prefill on the first post-compact call, then as
cached input until the next compaction (or end of trajectory).

\paragraph{Per-question token totals.} Table~\ref{tab:tokens_all}
reports, for each (model, benchmark, method) cell, the cumulative prompt
tokens $N_{\text{prompt}}$ and completion tokens $N_{\text{out}}$ summed
across all LLM calls (assistant turns, rubric probes, summarizer calls),
together with the split of $N_{\text{prompt}}$ into first-time prefill
$N_{\text{prefill}}$ and cache reads $N_{\text{cached}}$. The single-rate
cost reported in the main text (Table~\ref{tab:perf-cost-multi})
recovers directly from the $N_{\text{prompt}}$ column as
$(p_{\text{cache}}\,N_{\text{prompt}} + p_{\text{out}}\,N_{\text{out}})/10^{6}$.
The \textbf{Cost} column of Table~\ref{tab:tokens_all} instead applies
the finer two-rate billing,
$(p_{\text{in}}\,N_{\text{prefill}} + p_{\text{cache}}\,N_{\text{cached}}
+ p_{\text{out}}\,N_{\text{out}})/10^{6}$,
charging the first-time prefill at the higher uncached rate
$p_{\text{in}}$; this raises each entry modestly above the single-rate
value in Table~\ref{tab:perf-cost-multi} but preserves the headline
comparison that \method{} costs less than the no-compaction baseline.

  \begin{table}[h!]
  \centering
  \small
  \caption{Per-question token accounting across all three benchmarks, splitting the
  cumulative prompt $N_{\text{prompt}}$ into first-time prefill
  ($N_{\text{prefill}}$) and cache reads ($N_{\text{cached}}$), alongside output
  ($N_{\text{out}}$). Cost uses the two-rate billing
  $N_{\text{prefill}}\,p_{\text{in}} + N_{\text{cached}}\,p_{\text{cache}} + N_{\text{out}}\,p_{\text{out}}$
  with the prices in Table~\ref{tab:openrouter_pricing}; this is finer than the
  single-rate approximation behind Table~\ref{tab:perf-cost-multi} and so gives
  slightly higher costs.}
  \label{tab:tokens_all}
  \setlength{\tabcolsep}{4pt}
  \begin{tabular}{lllrrrrr}
  \toprule
  \textbf{Model} & \textbf{Benchmark} & \textbf{Method} & $N_{\text{prompt}}$ & $N_{\text{prefill}}$
  & $N_{\text{cached}}$ & $N_{\text{out}}$ & \textbf{Cost} \\
  \midrule
  \multirow{9}{*}{\texttt{GLM-4.7-Flash}}
   & \multirow{3}{*}{BrowseComp-Plus}
     & No Compaction  & 11.6M & 0.58M & 11.0M & 8.9k & \$0.15 \\
   & & Fixed-interval &  3.6M & 0.29M &  3.31M & 9k   & \$0.06 \\
   & & \method{}      &  3.6M & 0.29M &  3.31M & 9k   & \$0.06 \\
  \cmidrule(lr){2-8}
   & \multirow{3}{*}{BrowseComp}
     & No Compaction  & 13.6M & 0.68M & 12.9M & 8.9k & \$0.18 \\
   & & Fixed-interval &  5.6M & 0.45M &  5.15M & 9k   & \$0.09 \\
   & & \method{}      &  9.6M & 0.77M &  8.83M & 9k   & \$0.15 \\
  \cmidrule(lr){2-8}
   & \multirow{3}{*}{DeepSearchQA}
     & No Compaction  & 12.6M & 0.63M & 12.0M & 8.9k & \$0.17 \\
   & & Fixed-interval &  4.6M & 0.37M &  4.23M & 9k   & \$0.07 \\
   & & \method{}      &  6.6M & 0.53M &  6.07M & 9k   & \$0.10 \\
  \midrule
  \multirow{9}{*}{\texttt{MiniMax-M2.5}}
   & \multirow{3}{*}{BrowseComp-Plus}
     & No Compaction  &  5.7M & 0.29M &  5.41M & 16.6k & \$0.27 \\
   & & Fixed-interval &  0.7M & 0.06M &  0.64M & 16.9k & \$0.06 \\
   & & \method{}      &  1.7M & 0.14M &  1.56M & 17.4k & \$0.11 \\
  \cmidrule(lr){2-8}
   & \multirow{3}{*}{BrowseComp}
     & No Compaction  &  5.7M & 0.29M &  5.41M & 16.6k & \$0.27 \\
   & & Fixed-interval &  1.7M & 0.13M &  1.57M & 16.9k & \$0.11 \\
   & & \method{}      &  2.3M & 0.18M &  2.12M & 17.4k & \$0.14 \\
  \cmidrule(lr){2-8}
   & \multirow{3}{*}{DeepSearchQA}
     & No Compaction  &  5.7M & 0.29M &  5.41M & 16.6k & \$0.27 \\
   & & Fixed-interval &  1.3M & 0.11M &  1.19M & 16.9k & \$0.09 \\
   & & \method{}      &  2.0M & 0.16M &  1.84M & 17.4k & \$0.12 \\
  \midrule
  \multirow{9}{*}{\texttt{MiMo-V2-Flash}}
   & \multirow{3}{*}{BrowseComp-Plus}
     & No Compaction  & 23.8M & 1.19M & 22.6M & 6.8k & \$0.35 \\
   & & Fixed-interval & 13.8M & 1.10M & 12.7M & 7.0k & \$0.24 \\
   & & \method{}      & 15.8M & 1.26M & 14.5M & 7.2k & \$0.27 \\
  \cmidrule(lr){2-8}
   & \multirow{3}{*}{BrowseComp}
     & No Compaction  & 26.8M & 1.34M & 25.5M & 6.8k & \$0.39 \\
   & & Fixed-interval & 13.8M & 1.10M & 12.7M & 7.0k & \$0.24 \\
   & & \method{}      & 10.8M & 0.86M &  9.94M & 7.2k & \$0.19 \\
  \cmidrule(lr){2-8}
   & \multirow{3}{*}{DeepSearchQA}
     & No Compaction  & 24.8M & 1.24M & 23.6M & 6.8k & \$0.36 \\
   & & Fixed-interval & 13.8M & 1.10M & 12.7M & 7.0k & \$0.24 \\
   & & \method{}      & 12.8M & 1.02M & 11.8M & 7.2k & \$0.22 \\
  \bottomrule
  \end{tabular}
  \end{table}
\begin{table}[h!]
\centering
\small
\caption{Per-question token consumption for the math benchmarks
(IMO-Answerbench / HMMT~Nov~25 / HMMT~Feb~26 averaged). Prompt is the
cumulative refinement summaries fed back into the next round; output is
model generation per round. The Output column matches the
\texttt{[Xk]} tag in Table~\ref{tab:performance_fixed_vs_adaptive};
Total is Prompt + Output.}
\label{tab:tokens_math}
\setlength{\tabcolsep}{6pt}
\begin{tabular}{lrrr}
\toprule
\textbf{Method} & \textbf{Prompt (cum.)} & \textbf{Output} & \textbf{Total} \\
\midrule
\texttt{Qwen3-4B-Instruct-2507} \\[2pt]
\quad No Compaction        & 1.2k  &  16.0k &  17k \\
\quad Fixed-Interval       & 7.1k  &  44.1k &  51k \\
\quad \method{} & 7.9k  &  47.9k &  56k \\
\midrule
\texttt{Qwen3-30B-A3B-Instruct-2507} \\[2pt]
\quad No Compaction        & 1.2k  &  16.0k &  17k \\
\quad Fixed-Interval       & 4.1k  &  25.6k &  30k \\
\quad \method{}  & 4.6k  &  28.6k &  33k \\
\midrule
\texttt{Qwen3.5-9B (Thinking Disabled)} \\[2pt]
\quad No Compaction        & 1.2k  &  16.0k &  17k \\
\quad Fixed-Interval       & 14.4k &  89.0k & 103k \\
\quad \method{}  & 15.0k &  92.5k & 107k \\
\midrule
\texttt{Qwen3.5-4B (Thinking Disabled)} \\[2pt]
\quad No Compaction        & 1.2k  &  16.0k &  17k \\
\quad Fixed-Interval       & 10.1k &  63.7k &  74k \\
\quad \method{} & 10.7k &  66.6k &  78k \\
\bottomrule
\end{tabular}
\end{table}


\clearpage
%
%
\newtcolorbox{casebox}[1][]{
   enhanced, breakable,
   colback=orange!4,
   colframe=orange!75!black,
   colbacktitle=orange!75!black,
   coltitle=white,
   fonttitle=\bfseries,
   boxrule=0.8pt, arc=3pt,
   left=8pt, right=8pt, top=8pt, bottom=8pt,
   #1
}
 
\section{Qualitative trajectories}
\label{appendix:case_studies}
 
We walk through three BrowseComp-Plus questions on \texttt{MiniMax-M2.5},
each run two ways: \textbf{fixed-interval} compaction (summarize whenever the
prompt passes $30\%$ of the context window) and \textbf{\method{}}
(summarize only when the rubric judges that a sub-task has closed). On all
three questions both fixed-interval \emph{and} the no-compaction baseline
answer incorrectly, while \method{} answers correctly --- so the difference
is \emph{when} compaction fires, not whether it fires.
 
\paragraph{How to read these traces.} Each turn is one \texttt{search} or
\texttt{get\_document} call. Fixed-interval compacts on a blind clock: every
time the running prompt refills to ${\approx}60$k tokens ($30\%$ of
MiniMax-M2.5's window) it summarizes whatever is currently in context ---
including dead-end searches. \method{} probes a rubric at intervals and
compacts only when a sub-task has genuinely resolved, so it can let the
prompt grow well past $30\%$ while the agent is still exploring. We write
each compaction as (prefix tokens $\to$ summary tokens).
 
\begin{casebox}[title={Case~A: Whitesnake --- fixed-interval stays anchored, \method{} breaks out}]
\textbf{Question.} Name the band formed between 1970 and 1990 by a musician
who played in a rock band that sold 100M+ records, began performing as a
teenager, was married three times, has one daughter and one son, attended art
college, and worked in a boutique. \textbf{Gold:} \emph{Whitesnake} (the
musician is David Coverdale, ex-Deep Purple).
 
\textbf{Fixed-interval @ $30\%$ (wrong: ``Paul McCartney formed Wings'').}
\begin{itemize}[leftmargin=*]\setlength\itemsep{1pt}
\item \textbf{Turns 1--20.} Cycles famous British rockers against the clues ---
Keith Richards, Pete Townshend, Malcolm McLaren, Bryan Ferry --- none of whom fit.
\item \textbf{Compactions at turns 21, 41, 61, 82} (each ${\approx}60\mathrm{k}\to1$--$3\mathrm{k}$).
Each summary re-dumps the same ruled-out-names search log, so the agent
resumes from the identical dead-end shortlist.
\item \textbf{Turns 22--100.} Keeps re-trying the same pool (Brian May, Mick
Jagger, Ronnie Wood, Paul McCartney), never tests David Coverdale, hits the
100-search cap, and falls back to a generic guess.
\end{itemize}
\textbf{\method{} (correct: \emph{Whitesnake}).}
\begin{itemize}[leftmargin=*]\setlength\itemsep{1pt}
\item \textbf{Turns 1--40.} The same broad cycling (Freddie Mercury, Bowie,
Boy George, Stevie Nicks); the rubric keeps continuing --- no sub-task has
closed --- and lets the prompt grow to ${\approx}118$k.
\item \textbf{One compaction at turn 41} ($118.6\mathrm{k}\to1.1\mathrm{k}$),
distilling the constraints rather than the ruled-out shortlist.
\item \textbf{Turns 46--57.} No longer re-primed with the old names, the agent
tries a fresh candidate --- David Coverdale --- confirms Deep Purple (100M+
records), art college, and the boutique job, and answers \emph{Whitesnake}.
\end{itemize}
Fixed-interval compacted four times but kept re-seeding the same wrong shortlist;
\method{}'s single late compaction cleared it and let the agent escape to the
right musician.
\end{casebox}

 \clearpage
\begin{casebox}[title={Case~B: Majida El Roumi --- fixed-interval freezes a wrong lead}]
\textbf{Question.} An artist released their first solo album between 1975 and
1978, served as a UN Goodwill Ambassador, and made a song (released 2004--2007)
that is a rendition of a 1956--1959 classical piece; a different artist --- who
started singing at age eight and cited Barbra Streisand and Queen --- released a
version between 1994 and 1997. Name the first artist. \textbf{Gold:}
\emph{Majida El Roumi} (her ``Habibi'' is built on Albinoni's \emph{Adagio}).
 
\textbf{Fixed-interval @ $30\%$ (wrong: ``None'').}
\begin{itemize}[leftmargin=*]\setlength\itemsep{1pt}
\item \textbf{Turns 1--20.} Chases both threads but latches onto the wrong
classical piece --- Rachmaninoff, via C\'eline Dion's ``All by Myself''.
\item \textbf{Compactions at turns 21, 41, 61, 81} (${\approx}60\mathrm{k}$ each).
Every summary carries the ``All by Myself'' lead forward, so the agent stays on
Western pop (Nana Mouskouri, Eric Carmen, Cyndi Lauper, Ang\'elique Kidjo).
\item \textbf{Turns 22--100.} Never reconsiders the piece or tests an Arabic
artist; hits the cap and returns ``None''.
\end{itemize}
\textbf{\method{} (correct: \emph{Majida El Roumi}).}
\begin{itemize}[leftmargin=*]\setlength\itemsep{1pt}
\item \textbf{Turns 1--23.} Searches both threads and at turns 18--20 corrects
the piece --- Albinoni's \emph{Adagio}, not Rachmaninoff --- then looks for an
Adagio cover by a UN-Ambassador artist.
\item \textbf{One compaction at turn 24} ($70.2\mathrm{k}\to0.9\mathrm{k}$),
consolidating the corrected lead.
\item \textbf{Turns 25--42.} Zeroes in on Majida El Roumi and her song
``Habibi'', confirms the album dates, and answers correctly.
\end{itemize}
The fixed clock froze an early wrong guess into every summary; \method{}
compacted only after the agent had corrected the lead to Albinoni's \emph{Adagio}.
\end{casebox}
 
\begin{casebox}[title={Case~C: Raheem Sterling --- fixed-interval never reaches the match}]
\textbf{Question.} A league football match (1992--2023) in a European capital,
attendance 61{,}700--61{,}906; Team Y scored in the 6th minute (a left-footed
player), Team X scored in the 35th and 75th minutes and in stoppage time
(90'+4', 90'+7'). Name the player who assisted the third goal (the
75th-minute one). \textbf{Gold:} \emph{Raheem Sterling} (Tottenham 1--4
Chelsea, 6 Nov 2023).
 
\textbf{Fixed-interval @ $30\%$ (wrong: ``Unable to determine'').}
\begin{itemize}[leftmargin=*]\setlength\itemsep{1pt}
\item \textbf{Turns 1--22.} Searches the goal-minute fingerprint
(6'/35'/75'/90+4'/90+7') and the attendance band; the literal-minute queries
return nothing relevant. Guesses Sevilla--Roma.
\item \textbf{Compactions at turns 23, 45, 67, 88} (${\approx}60\mathrm{k}$ each).
Each summary re-dumps the same fruitless minute-by-minute log, and the agent
keeps trying stadiums and teams at random (Stadio Olimpico, PSG, Man Utd, Moscow).
\item \textbf{Turns 24--100.} Never tests Tottenham--Chelsea; hits the cap and
gives up.
\end{itemize}
\textbf{\method{} (correct: \emph{Raheem Sterling}).}
\begin{itemize}[leftmargin=*]\setlength\itemsep{1pt}
\item \textbf{Turns 1--22.} The same fingerprint searches, also unproductive.
\item \textbf{Turns 23--25.} Tries the right pairing --- Tottenham vs Chelsea,
4--1, November 2023 --- and hits the match.
\item \textbf{One compaction at turn 28} ($70\mathrm{k}\to0.8\mathrm{k}$),
consolidating the identified match; turns 29--41 confirm the attendance,
Kulusevski's left-footed 6th-minute goal, and Sterling's assist on the
75th-minute goal.
\end{itemize}
Both arms flailed on the goal-minute fingerprint, but \method{} kept exploring
until the match was found and compacted only then; the fixed clock kept
interrupting to re-summarize dead ends and never reached the pairing.
\end{casebox}

\clearpage
\end{document}